\documentclass[letterpaper, 10 pt, conference]{ieeeconf}  %

\IEEEoverridecommandlockouts                              %

\usepackage[utf8]{inputenc} %
\usepackage[T1]{fontenc}    %
\usepackage{hyperref}       %
\usepackage{url}            %
\usepackage{booktabs}       %
\usepackage{amsfonts}       %
\usepackage{nicefrac}       %
\usepackage{microtype}      %
\usepackage{xcolor}         %

\usepackage{cite}
\usepackage{flushend}

\usepackage{amsmath}
\usepackage{amssymb}
\usepackage{mathtools}

\usepackage{color}

\usepackage{amsfonts}
\usepackage{booktabs}
\usepackage{algorithm}
\usepackage{algorithmicx}
\usepackage[noend]{algpseudocode}
\usepackage{wrapfig}
\usepackage{bm}
\usepackage{bbm}
\usepackage{breqn}
\usepackage[font=footnotesize]{caption}
\usepackage{array}
\usepackage{multirow}
\usepackage{csquotes}
\usepackage{adjustbox}

\newcommand{\bx}{\mathbf{x}}
\newcommand{\bs}{\mathbf{s}}
\newcommand{\ba}{\mathbf{a}}
\newcommand{\bz}{\mathbf{z}}

\DeclareMathOperator*{\argmin}{arg\,min}

\DeclarePairedDelimiterX{\kldivx}[2]{(}{)}{%
  #1\;\delimsize\|\;#2%
}
\newcommand{\kldiv}{D_{\mathrm{KL}}\kldivx}

\title{\LARGE \bf
Bootstrapping Adaptive Human-Machine Interfaces\\with Offline Reinforcement Learning
}

\author{Jensen Gao$^{1, 2}$, Siddharth Reddy$^2$, Glen Berseth$^{2,3,4}$, Anca D. Dragan$^2$, Sergey Levine$^2$%
\thanks{$^1$Stanford University, $^2$University of California, Berkeley, $^3$Universit\'e de Montr\'eal, $^4$MILA. Contact: \texttt{jenseng@stanford.edu}. Appendix available at \url{https://sites.google.com/view/orbit-assist}.}%
        }
        
\begin{document}

\maketitle
\thispagestyle{empty}
\pagestyle{empty}

\begin{abstract}
Adaptive interfaces can help users perform sequential decision-making tasks like robotic teleoperation given noisy, high-dimensional command signals (e.g., from a brain-computer interface).
Recent advances in human-in-the-loop machine learning enable such systems to improve by interacting with users, but tend to be limited by the amount of data that they can collect from individual users in practice.
In this paper, we propose a reinforcement learning algorithm to address this by training an interface to map raw command signals to actions using a combination of offline pre-training and online fine-tuning.
To address the challenges posed by noisy command signals and sparse rewards, we develop a novel method for representing and inferring the user's long-term intent for a given trajectory.
We primarily evaluate our method's ability to assist users who can only communicate through noisy, high-dimensional input channels through a user study in which 12 participants performed a simulated navigation task by using their eye gaze to modulate a 128-dimensional command signal from their webcam.
The results show that our method enables successful goal navigation more often than a baseline directional interface, by learning to denoise user commands signals and provide shared autonomy assistance.
We further evaluate on a simulated Sawyer pushing task with eye gaze control, and the Lunar Lander game with simulated user commands, and find that our method improves over baseline interfaces in these domains as well. 
Extensive ablation experiments with simulated user commands empirically motivate each component of our method.

\end{abstract}

\section{Introduction} \label{intro}

\begin{figure}[t]
  \begin{center}
    \includegraphics[width=\linewidth]{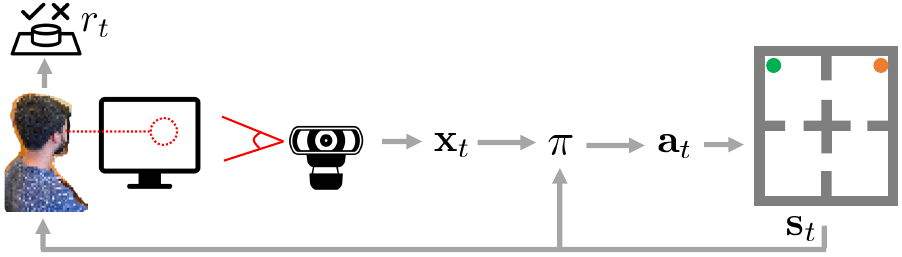}
    \caption{The user provides a noisy, high-dimensional command signal $\bx_t$, the interface $\pi$ takes an action $\ba_t$, and the user provides reward feedback $r_t$.}
    \label{fig:data-collection}
    \vspace{-25pt}
  \end{center}
\end{figure}

One of the central problems in the field of human-computer interaction is designing interfaces that help users control complex systems, such as prosthetic limbs and assistive robots, by translating raw user commands (e.g., brain signals) into actions (illustrated in Fig. \ref{fig:data-collection}).
Recent work proposes various human-in-the-loop reinforcement learning (RL) algorithms that address this challenge by training the interface to maximize positive user feedback on the system's performance \cite{pilarski2011online,broad2017learning,reddy2018shared,schaff2020residual,du2020ave,x2t2021,asha2022}.
This approach customizes the interface to individual users and improves with use, but tends to require a large amount of human interaction data.
Prior work improves data efficiency by assuming access to some combination of the user's task distribution, a task-agnostic reward function, and a prior mapping from raw user commands into actions \cite{reddy2018shared,x2t2021,asha2022}, which can be difficult to specify in practice.
In this paper, we lift these assumptions using recent advances in offline RL \cite{fujimoto2019off,wu2019behavior,kumar2020conservative,yu2020mopo,kidambi2020morel,nair2020awac,shen2021model,emmons2021rvs}, to instead leverage offline data for improved online data efficiency.
Instead of assuming prior knowledge of the user's desired tasks, we take a purely data-driven approach to interface optimization.
We propose an offline RL algorithm for interface optimization that can learn from both an observational dataset of the user attempting to perform their desired tasks using some unknown existing default interface, as well as online data collected using our learned interface, where each episode is labeled with a sparse reward that indicates overall task success or failure. We assume that the default interface is performant enough for the user to produce data that contains useful behavior to learn from. 
To address the high cost of user data, we pre-train value functions on a large offline dataset, and distill them into a policy (i.e., an interface) that infers the user's desired action from the user's command signal.
To enable the user and interface to co-adapt and discover command styles that are different from those in the offline data, we then fine-tune the interface through online, human-in-the-loop RL.
In our experiments, we instantiate this approach with implicit Q-learning (IQL) \cite{kostrikov2021offline}.
We call our method the \emph{Offline RL-Bootstrapped InTerface} (ORBIT). \looseness=-1

There are several challenging features of our problem setting that resist straightforward applications of existing offline RL algorithms.
First, the user's command is noisy and only provides a partial observation of the user's long-term intent.
Second, we assume that users can only provide sparse, terminal rewards, as more complex forms of reward feedback can be difficult to specify in practice.
These factors, combined with practical limitations on the size of the offline dataset, makes learning a representation of the user's intent challenging.
As such, we train representations of user intent to be predictive of the user's rewards, which we use to condition our offline RL algorithm on.
To avoid overfitting,
we propose to regularize these learned representations using negative data augmentation and a variational information bottleneck.

Our setting also differs from the standard offline RL regime in that, during online fine-tuning, the user can adapt to the interface to operate it more effectively.
While this can make RL more difficult due to non-stationary dynamics, it also presents an opportunity to learn an interface that not only denoises commands and autocompletes actions, but departs from the overall command strategy of the default interface and the offline data.

\begin{figure*}[t]
    \centering
    \vspace{0.05in}
    \includegraphics[width=\linewidth]{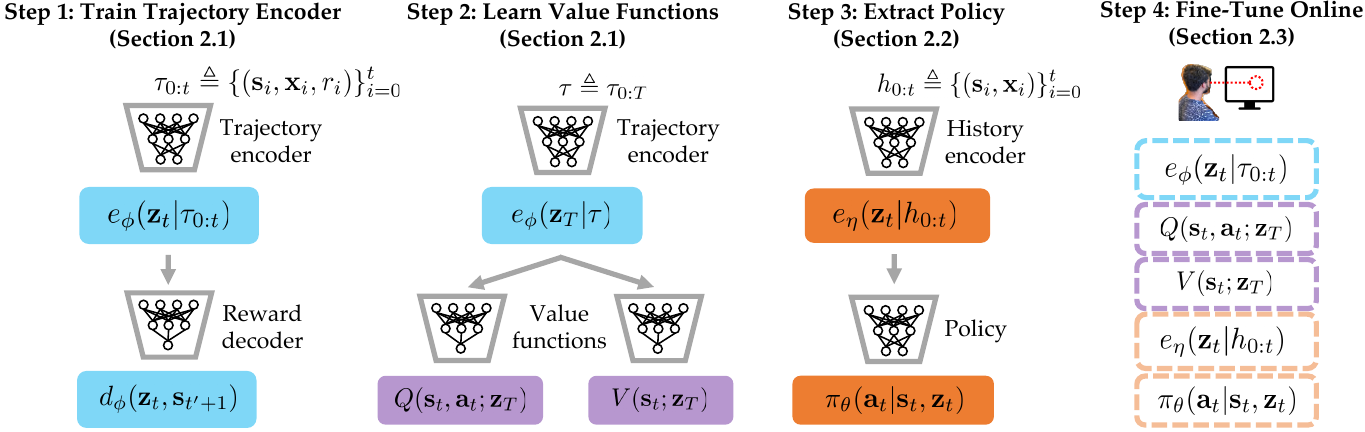}
    \caption{An overview of our ORBIT method and model architectures. In steps 1 and 2, we learn value functions $Q$ and $V$ (purple) from the offline data using implicit Q-learning \cite{kostrikov2021offline}. Since the user's intent is only partially observable given their commands $\bx_i$, we condition the predicted values $Q(\bs_t, \ba_t; \bz_T)$ and $V(\bs_t; \bz_T)$ on a latent embedding $\bz_T$ of the full trajectory $\tau$ that the state $\bs_t$ came from (Sec. \ref{learn-value}).
    In step 3, we use the learned value functions $Q$ and $V$ to extract a policy $\pi$ (Sec. \ref{extract-policy}).
    After pre-training the value functions $Q$ and $V$ and policy $\pi$ in this manner, we efficiently fine-tune them through online RL with the user in the loop in step 4 (Sec. \ref{fine-tuning}).}
    \label{fig:schematic}
    \vspace{-15pt}
\end{figure*}

To our knowledge, ORBIT is the first human-in-the-loop RL algorithm that can learn an interface from both offline and online data.
One key improvement over prior work is a novel representation learning method for encoding and conditioning on full trajectories in the value function (Sec. \ref{learn-value}), and for conditioning on the history of previous states and commands in the policy (Sec. \ref{extract-policy}).
To evaluate ORBIT's ability to learn an effective interface from real user data, we conduct a user study with 12 participants who perform a simulated navigation task by using their eye gaze to modulate a 128-dimensional command signal from their webcam (illustrated in Fig. \ref{fig:nav-setup} and \ref{fig:nav-heat}).
The results show that ORBIT enables the user to successfully navigate to their goal more often than a default, directional interface used to collect the training data. We also showcase ORBIT on a simulated Sawyer pushing task with input from a single human user, and on the Lunar Lander game with simulated user input (see Table \ref{tab:sawyer-lander-results}).

\section{Interface Optimization via Offline RL} \label{methods}

In our setting, the user cannot directly take actions, and relies on an interface to convert the user's raw command signals into actions.
The interface cannot directly observe any specification of the desired task.
We formulate the assistance problem as a contextual Markov decision process (CMDP) \cite{hallak2015contextual}.
Each observation consists of two variables: the state of the environment $\bs_t$ (e.g., the position of the robot) and the user's command $\bx_t$ (e.g., brain signals).
The interface $\pi(\ba_t | \bs_{0:t}, \bx_{0:t})$ takes an action $\ba_t$ given a history of previous states $\bs_{0:t}$ and commands $\bx_{0:t}$.
We do not assume that we can observe the user's desired task (which defines the context in the CMDP), the space of possible tasks, or the ground-truth reward function $R(\bs_t, \ba_t)$.
Instead, we assume access to a reward signal $r_t$ that comes from the user---to simplify our method and experiments, we assume that the user only provides a binary reward $r_T \in \{0, 1\}$ at the end of each episode to indicate whether the interface succeeded or failed at the user's desired task, and automatically set non-terminal rewards $r_{t < T}$ to zero.
We approach this problem by using offline RL and online fine-tuning to train an interface $\pi$ to maximize the user-provided rewards.
Fig. \ref{fig:schematic} outlines our method. \looseness=-1

\subsection{Learning Value Functions from Partial Observations} \label{learn-value}

In order to train an effective interface, we would like to learn the action-value function $Q$ and state-value function $V$ from an offline dataset of trajectories in which the user gives commands, some unknown default interface executes actions, and the user gives reward feedback.
However, the user's commands $\bx_t$ only provide partial observations of the user's long-term intent.
To accurately model the value function, we represent the user's intent as a function of the full sequence of observed environment states and corresponding commands.
To accomplish this, we learn a trajectory encoder $e_{\phi}$ that takes in a sequence of tuples $\{(\bs_i, \bx_i, r_i)\}_{i=0}^t$, denoted as a partial trajectory $\tau_{0:t}$, and outputs a distribution over the latent embedding $\bz_t$ that represents the user's intent (see Step 1 in Fig. \ref{fig:schematic}).
Note that this trajectory encoder also takes rewards as input, since rewards contain important information about the user's intent in retrospect.
We then condition both value functions $Q(\bs_t, \ba_t; e_{\phi}(\tau))$ and $V(\bs_t; e_{\phi}(\tau))$ on a latent embedding $e_{\phi}(\tau)$ of the full trajectory $\tau$ that the input $(\bs_t, \ba_t)$ comes from, where $e_{\phi}(\tau)$ denotes the mean embedding $\mathbb{E}_{\bz_t \sim e_{\phi}(\bz_t | \tau)}[\bz_t | \tau]$. \looseness=-1

To ensure that the latents $\bz_t$ contain information about the user's intent, we train the trajectory encoder such that the latents are predictive of the user's reward labels, akin to prior work \cite{zintgraf2019varibad}.
In particular, we train $e_{\phi}$ end to end with a reward decoder $d_{\psi}$ that predicts the reward $r_{t'}$ given the pair $(\bz_t, \bs_{t'+1})$ for any $t$ (including $t \neq t'$).
For the reward decoder to accurately predict the reward $r_{t'}$ for entering the state $\bs_{t'+1}$, the latent $\bz_t$ must contain information about the user's desired task inferred from the partial trajectory $\tau_{0:t}$.
We jointly train the trajectory encoder $e_{\phi}$ and the reward decoder $d_{\psi}$ to minimize the reward prediction error,
\vspace{-0.05in}
\begin{equation} \label{eqn:rpe}
\small
\ell_{\mathrm{RPE}}(\phi, \psi) \triangleq \sum_{\tau \in \mathcal{D}} \sum_{t=0}^T \sum_{t'=0}^T H(d_{\psi}(e_{\phi}(\tau_{0:t}), \bs_{t'+1}), r_{t'}),
\end{equation}
where $H$ is the cross entropy, and $\mathcal{D}$ is an offline dataset of trajectories.
The ablation experiment for \textbf{Q8} in Sec. \ref{ablations} shows that training the trajectory encoder to predict rewards performs much better than end to end with the value functions. \looseness=-1

The purpose of the reward decoder is to force the latent $\bz_t$ to contain useful information about the user's intent, but without any regularization, it learns to accurately predict the reward $r_{t'}$ based on the state $\bs_{t'+1}$ alone while ignoring the latent $\bz_t$, causing the trajectory encoder $e_{\phi}$ to generate uninformative latents.
The ablation experiment for \textbf{Q4} in Sec. \ref{ablations} shows that this effect substantially degrades the performance of the interface.
To encourage the reward decoder to use the latent $\bz_t$ to predict the reward $r_{t'}$, we also train on an auxiliary loss term that mimics the binary cross-entropy loss in Eqn. \ref{eqn:rpe}, but pairs the states $\bs_{t'+1}$ from trajectory $\tau$ with the latents $\bz_t$ from a different trajectory $\tau'$, and labels the resulting pairs $(\bz_t, \bs_{t'+1})$ with zero rewards $r_{t'} = 0$.
We refer to this term as the negative data augmentation loss,
\vspace{-0.05in}
\begin{equation} \label{eqn:nda}
\small
\ell_{\mathrm{NDA}}(\phi, \psi) \triangleq \sum_{\tau \in \mathcal{D}} \sum_{t=0}^T \sum_{t'=0}^T - \log{(1 - d_{\psi}(e_{\phi}(\tau'_{0:t}), \bs_{t'+1}))},
\end{equation}
where the trajectory $\tau'$ is randomly sampled from the dataset $\mathcal{D}$, and $s_{t'+1}$ belongs to the trajectory $\tau$.
To prevent the reward decoder from overfitting to embeddings of the offline trajectories, we regularize the upstream trajectory encoder with a variational information bottleneck (VIB) \cite{alemi2016deep,achille2018information},
\vspace{-0.05in}
\begin{equation} \label{eqn:vib}
\small
\ell_{\mathrm{VIB}}(\phi) \triangleq \sum_{\tau \in \mathcal{D}} \sum_{t=0}^T  \kldiv{e_{\phi}(\bz_t | \tau_{0:t})}{\mathcal{N}(\mathbf{0}, I_k)},
\end{equation}
where $k$ is the dimensionality of the latent space.
The ablation experiment for \textbf{Q5} in Sec. \ref{ablations} shows that this VIB is critical to the empirical performance of the interface.

Putting the reward prediction error, negative data augmentation, and information bottleneck losses together, we have
\vspace{-0.05in}
\begin{equation} \label{eqn:val-seq-enc-loss}
\small
\ell(\phi, \psi) \triangleq \ell_{\mathrm{RPE}}(\phi, \psi) + \ell_{\mathrm{NDA}}(\phi, \psi) + \beta_1 \ell_{\mathrm{VIB}}(\phi),
\end{equation}
where $\beta_1$ is a constant hyperparameter.
In our experiments, we model the trajectory encoder $e_{\phi}$ as a gated recurrent neural network (RNN) \cite{cho2014properties}, and the reward decoder $d_{\psi}$ as a feedforward neural network.
After training $e_{\phi}$ to minimize the loss in Eqn. \ref{eqn:val-seq-enc-loss}, we fit the value functions $Q(\bs_t, \ba_t; e_{\phi}(\tau))$ and $V(\bs_t; e_{\phi}(\tau))$ by optimizing the value losses $L_Q$ and $L_V$ from implicit Q-learning (IQL) \cite{kostrikov2021offline}.\

While we intend for the trajectory encoder $e_{\phi}$ to help address the challenges of estimating user intent from sparse rewards, it does not directly address the more general exploration and optimization challenges of RL with sparse rewards, which is outside the scope of this work. \looseness=-1

\subsection{Extracting a Policy from the Value Functions} \label{extract-policy}

We would like to train a policy to maximize the learned values.
Due to the partial observability of the user's long-term intent given their commands, this policy should be conditioned on the history of previous states and commands.
We could simply reuse the learned trajectory encoder $e_{\phi}$ from the previous section to embed this history, along with rewards set to zero, as $e_{\phi}$ is trained to produce latents informative of user intent from this input. However, the ablation experiment for \textbf{Q6} in Sec. \ref{ablations} shows that this performs substantially worse than baseline methods. \looseness=-1

Instead, we train a new `history encoder' $e_{\eta}$ that takes a history $h_{0:t} \triangleq \{(\bs_i, \bx_i)\}_{i=0}^t$ as input, and outputs a distribution over the latent embedding $\bz_t$. 
Using this history encoder, we model the policy as $\pi_{\theta}(\ba_t | \bs_t; e_{\eta}(h_{0:t}))$, where $e_{\eta}(h_{0:t})$ denotes the mean embedding $\mathbb{E}_{\bz_t \sim e_{\eta}(\bz_t | h_{0:t})}[\bz_t | h_{0:t}]$.
We train the policy and history encoder on the weighted behavioral cloning (BC) loss from \cite{peng2019advantage},
\vspace{-0.05in}
\begin{equation} \label{eqn:weighted-bc}
\small
\ell_{\mathrm{WBC}}(\theta, \eta) \triangleq \sum_{\tau \in \mathcal{D}} \sum_{t=0}^T -\log{\pi_{\theta}(\ba_t | \bs_t; e_{\eta}(h_{0:t}))}e^{\beta_2 A(\bs_t, \ba_t; e_{\phi}(\tau))},
\end{equation}
where $\beta_2$ is a constant hyperparameter, the weights are the exponentiated advantages $A(\bs_t, \ba_t; e_{\phi}(\tau)) \triangleq Q(\bs_t, \ba_t; e_{\phi}(\tau)) - V(\bs_t; e_{\phi}(\tau))$, and $Q$ and $V$ are the value functions learned through IQL in Sec. \ref{learn-value}.
Note that the policy is conditioned on the embedding $e_{\eta}(h_{0:t})$ from the history encoder $e_{\eta}$, while the advantages are obtained by conditioning on the full trajectory embedding $e_{\phi}(\tau)$ from the trajectory encoder $e_{\phi}$.
Step 3 in Fig. \ref{fig:schematic} illustrates this policy architecture in relation to our overall method.

As with the trajectory encoder, we regularize the history encoder using a VIB---the ablation experiment for \textbf{Q5} in Sec. \ref{ablations} shows that this is critical for performance.
Putting the weighted BC and VIB losses together, we have the policy extraction loss,
\vspace{-0.05in}
\begin{equation} \label{eqn:pol-loss}
\small
\ell(\theta, \eta) \triangleq \ell_{\mathrm{WBC}}(\theta, \eta) + \beta_3 \ell_{\mathrm{VIB}}(\eta),
\end{equation}
where $\beta_3$ is a constant hyperparameter, and $\ell_{\mathrm{VIB}}(\eta)$ is analogous to Eqn. \ref{eqn:vib}.
In our experiments, we model the policy $\pi_{\theta}$ as a feedforward network, and the history encoder $e_{\eta}$ as a permutation-invariant product of independent Gaussian factors similar to prior work \cite{rakelly2019efficient}.
We use this architecture for the history encoder, instead of an RNN like the trajectory encoder, because while the trajectory encoder is only evaluated in-distribution, the history encoder is evaluated during online rollouts, where there can be distribution shift. Thus, it benefits from the regularization of a permutation-invariant architecture.

In summary, the history encoder differs from the trajectory encoder in the following ways: \textbf{1.} It is used to condition the policy. \textbf{2.} It is only conditioned on past histories, because this is what is available during online rollouts. \textbf{3.} It is trained end to end with the weighted BC objective. \textbf{4.} It uses a permutation-invariant architecture for greater regularization.

\subsection{Online Fine-Tuning} \label{fine-tuning}

The previous sections describe how to pre-train a policy $\pi_{\theta}$
on an offline dataset $\mathcal{D}$.
However, pre-training alone may not outperform the default interface used to generate the offline data. 
To address this issue, we have the user operate the pre-trained interface, and further optimize it through online RL.
Fine-tuning the interface with the user in the loop generates on-policy training data and explores new interfaces.

During the online fine-tuning phase, we update the value functions, which first involves updating the learned representations $\bz_t$ of the user's intent.
To do so, we update the trajectory encoder $e_{\phi}$ by taking $n$ gradient steps on the loss in Eqn. \ref{eqn:val-seq-enc-loss} after each episode.
To reduce variance in the learning signal, we remove negative data augmentation---augmentation is not required because the decoder $d_{\psi}$ has already learned to use the latent $\bz_t$ for reward prediction during the offline learning phase.
To further stabilize fine-tuning of the trajectory encoder $e_{\phi}$, we also implement a trust region loss that keeps the fine-tuned trajectory encoder close to the pre-trained (and frozen) trajectory encoder: we freeze the reward decoder $d_{\psi}$ and pre-trained trajectory encoder $e_{\bar{\phi}}$, and minimize the KL divergence between the fine-tuned trajectory encoder $e_{\phi}$ and the pre-trained trajectory encoder $e_{\bar{\phi}}$ on the offline data $\mathcal{D}$,
\vspace{-0.05in}
\begin{equation} \label{eqn:trust-region}
\small
\ell_{\mathrm{TR}}(\phi) \triangleq \sum_{\tau \in \mathcal{D}} \sum_{t=0}^T \kldiv{e_{\bar{\phi}}(\bz_t | \tau_{0:t})}{e_{\phi}(\bz_t | \tau_{0:t})}.
\end{equation}
We then update the value functions $Q$ and $V$ by taking $n$ gradient steps on the value losses from IQL ($L_Q$ and $L_V$ in \cite{kostrikov2021offline}).
Finally, we update the policy $\pi_{\theta}$ and history encoder $e_{\eta}$ by taking $m$ gradient steps on the loss in Eqn. \ref{eqn:pol-loss}.
To prevent catastrophic forgetting, each mini-batch $\mathcal{B}$ of trajectories is balanced to consist of 50\% offline and 50\% online data. While the offline data may include lower quality samples, using offline RL should help address learning from suboptimal data.

\begin{algorithm}[ht]
\small
\begin{algorithmic}[1]
\State{$\phi, \psi \leftarrow \argmin_{\phi, \psi} \ell(\phi, \psi; \mathcal{D})$ \Comment{\emph{pre-train trajectory encoder and reward decoder (see Eqn. \ref{eqn:val-seq-enc-loss})}}}
\State{$Q, V \leftarrow \textproc{IQL}(\mathcal{D})$ \Comment{\emph{pre-train value functions through implicit Q-learning}}}
\State{$\theta, \eta \leftarrow \argmin_{\theta, \eta} \ell(\theta, \eta; \mathcal{D})$ \Comment{\emph{pre-train policy and history encoder (see Eqn. \ref{eqn:pol-loss})}}}
\State{$\bar{\phi} \leftarrow \phi$ \Comment{\emph{store frozen copy of trajectory encoder weights}}}
\State{$\mathcal{D}_{\text{online}} \leftarrow \emptyset$}
\For{each new episode $\tau$ of online fine-tuning where the user operates the learned interface $\pi_{\theta}$}
\State{$\mathcal{D}_{\text{online}} \leftarrow \mathcal{D}_{\text{online}} \cup \{\tau\}$}
\For{$j \in \{1, 2, ..., n\}$} \label{lst:line:n-grad}
\State{Sample mini-batch $\mathcal{B}$, half from $\mathcal{D}$ and half from $\mathcal{D}_{\text{online}}$}
\State{$\phi \leftarrow \phi - \nabla_{\phi} (\ell_{\mathrm{RPE}}(\phi, \psi; \mathcal{B}) + \beta_1 \ell_{\mathrm{VIB}}(\phi; \mathcal{B}) + \beta_4 \ell_{\mathrm{TR}}(\phi; \mathcal{D}, \bar{\phi}))$ \Comment{\emph{update traj. encoder (see Eqns. \ref{eqn:rpe}, \ref{eqn:vib}, \ref{eqn:trust-region})}}} \label{lst:line:fine-tune-traj-enc}
\State{Update value functions $Q$ and $V$ by taking 1 gradient step on $L_Q$ and $L_V$ from \textproc{IQL}}
\EndFor
\For{$j \in \{1, 2, ..., m\}$} \label{lst:line:m-grad}
\State{Sample mini-batch $\mathcal{B}$, half from $\mathcal{D}$ and half from $\mathcal{D}_{\text{online}}$}
\State{$\eta \leftarrow \eta - \nabla_{\eta} \ell(\theta, \eta; \mathcal{B})$ \Comment{\emph{update history encoder}}}
\State{$\theta \leftarrow \theta - \nabla_{\theta} \ell(\theta, \eta; \mathcal{B})$ \Comment{\emph{update policy (see Eqn. \ref{eqn:pol-loss})}}}
\EndFor
\EndFor
\end{algorithmic}
\caption{\textproc{ORBIT}($\mathcal{D}$)}
\label{alg:orbit-alg}
\end{algorithm}

Algorithm \ref{alg:orbit-alg} outlines the full method,
which we call the \emph{Offline RL-Bootstrapped InTerface} (ORBIT).
We first take an offline dataset $\mathcal{D}$ of trajectories labeled with sparse rewards, and learn a trajectory encoder $e_{\phi}$, value functions $Q$ and $V$, reward decoder $d_{\psi}$, history encoder $e_{\eta}$, and policy $\pi_{\theta}$ (see Fig. \ref{fig:schematic} for an overview).
We then fine-tune these models (except the reward decoder) on online trajectories collected with the user as they operate the learned interface $\pi_{\theta}$.

\section{Related Work}

The literature on learning-based assistive interfaces spans several fields, including brain-computer interfaces \cite{gilja2012high,dangi2013design,dangi2014continuous,merel2015neuroprosthetic,anumanchipalli2019speech,willett2020high}, natural language interfaces \cite{wang2016learning,karamcheti2020learning,ahn2022can}, speech interfaces \cite{fels1993glove,fels1997glove}, electronic musical instruments \cite{hunt2002mapping}, and robotic teleoperation interfaces \cite{kim2006continuous,mcmullen2013demonstration,carlson2012collaborative,argall2016modular,javdani2017acting,jeon2020shared}.
This prior work assumes access to some combination of a user model, candidate tasks, and ground-truth action labels for commands, which restricts them to settings with substantial prior knowledge or supervision.
Recent work on human-in-the-loop RL \cite{reddy2018shared,x2t2021,asha2022} lifts these assumptions, but still assumes access to either a task-agnostic reward function, task distribution, or prior interface.
In contrast, ORBIT only assumes access to a dataset of trajectories where the user provided a command, the system took an action, and the user provided a sparse reward signal. 
This purely data-driven approach to interface optimization makes ORBIT more practical for a broad range of real-world applications, where users can easily generate trajectory data but cannot manually specify their task distribution or a prior mapping. While the performance of ORBIT likely depends on the quality and diversity of the offline data, as this is common to data-driven learning approaches including offline RL, data of reasonable quality can likely be obtained in practice from existing default interfaces. 
Prior work on COACH \cite{macglashan2017interactive,arumugam2019deep}, TAMER \cite{knox2009interactively,warnell2017deep}, and preference learning \cite{dorsa2017active,christiano2017deep,ibarz2018reward,stiennon2020learning,biyik2021learning} trains RL agents from human feedback.
ORBIT differs in that it aims to train a user interface that can infer long-term intent from noisy, high-dimensional commands, rather than an autonomous agent in a fully-observable environment.

\begin{figure}[h]
  \begin{center}
    \includegraphics[width=\linewidth]{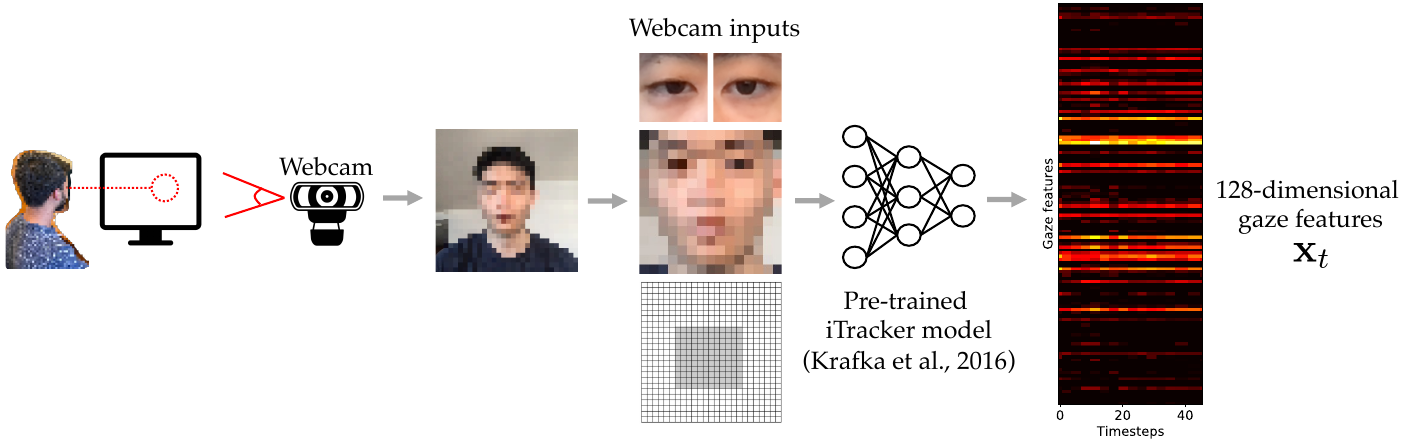}
    \caption{Command signals $\bx_t$ are obtained through the pipeline described in \cite{asha2022}, which records user webcam images and transforms them into a 128-dimensional latent embedding that represents eye gaze features. These features are predictive of the user's gaze position, but do not directly represent 2D locations, which makes inferring user intent from them challenging.}
    \label{fig:nav-setup}
    \vspace{-15pt}
  \end{center}
\end{figure}

\section{User Study} \label{user-study}

Our experiments focus on evaluating ORBIT's ability to learn an effective interface through a combination of offline pre-training and online fine-tuning.
To do so, we conduct a user study with 12 participants who use their eye gaze to modulate a noisy command signal from their webcam (see Fig. \ref{fig:nav-setup}) to perform the simulated 2D continuous navigation task from \cite{ghosh2019learning} (illustration in Fig. \ref{fig:nav-heat}a).
The state $\bs_t$ is the current 2D position, the user's command $\bx_t$ is a 128-dimensional signal from their webcam, and the action $\ba_t$ is a 2D velocity.
The eye gaze signals are the same used in prior work on RL-based adaptive interfaces \cite{x2t2021,asha2022}, and consist of representations from iTracker \cite{krafka2016eye}.

\begin{figure*}[t]
    \centering
    \vspace{0.05in}
    \includegraphics[width=\linewidth]{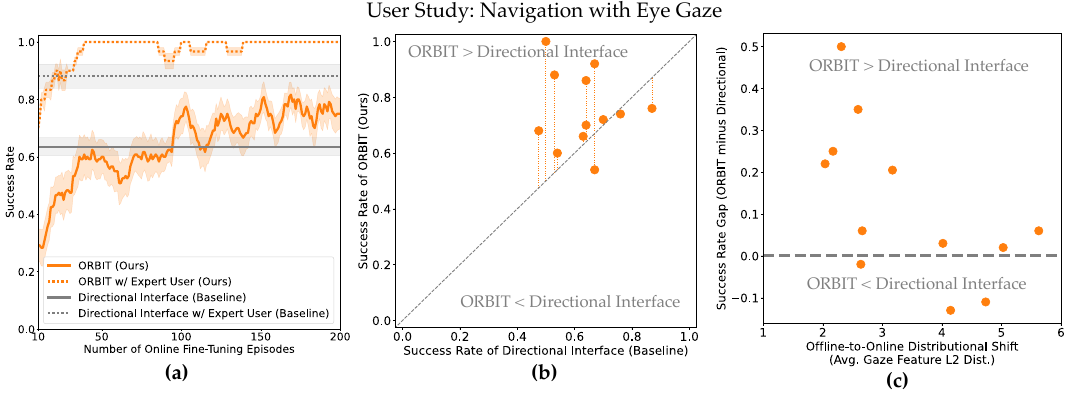}
    \caption{\textbf{(a)} Through offline pre-training and online fine-tuning, ORBIT (solid orange) learns an interface that outperforms the default interface (solid gray) that was used to generate the offline data. ORBIT performs even better when the same user who generated the offline data is brought back in for the online fine-tuning phase (dashed orange vs. gray). \textbf{(b)} ORBIT is most helpful for users who struggle with the directional interface. Each circle represents the success rate for one participant. \textbf{(c)} ORBIT works best when the online user is similar to the user who generated the offline data. Each circle represents the difference between one participant's success rate with ORBIT and their success rate with the directional interface. The x-axis represents the L2 distance between the mean command signal of the expert user's calibration data and those of each participant's.}
    \label{fig:nav-results}
    \vspace{-15pt}
\end{figure*}

While iTracker is no longer the state of the art in gaze tracking, our goal here is not to develop the best possible gaze control interface, but to study if RL can be used to learn a better control interface that operates on noisy command signals.
We also note that while this 2D navigation task is relatively simple, it represents a more challenging control task than those in prior work that use this command signal, which are limited to either contextual bandits, or settings where individual command signals are never mapped directly to low-level actions. \looseness=-1

\noindent\textbf{Baseline: directional interface.}
The `default' interface enables the user to move left/right/up/down by looking at the left/right/top/bottom half of the screen.
We calibrate this interface once at the start of each study by asking the user to look at the left, right, top, and bottom portions of the screen one by one, recording 20 webcam images for each location, and training a 2D gaze position estimator conditioned on iTracker command signals, as done in the original work \cite{krafka2016eye}.
In our setting, this calibrated interface is a strong baseline, since it assumes access to commands and ground-truth action labels for supervised calibration, whereas our ORBIT method does not assume access to paired data. While this interface may not be optimal for this task, we aim not to design the best possible baseline interface, but to investigate whether ORBIT can be used to improve upon the interface used to collect its training data. \looseness=-1

\noindent\textbf{Experiment design.}
Each of the 12 participants completed two phases of experiments.
In phase A, they use the default, directional interface for 100 episodes.
Note that the directional interface is non-adaptive, so it cannot improve with additional evaluation.
In phase B, they operate an adaptive interface $\pi_{\theta}$ that is initially pre-trained on the offline data then fine-tuned online by ORBIT for 200 episodes.
Each episode takes approximately 18 seconds on average.
To avoid the confounding effect of user improvement or fatigue, we counterbalance the order of phases A and B.
Users are informed of how each interface operates before each phase.

\noindent\textbf{Cold start problem.}
New users can face the ``cold start problem'' of not having past data for training their interface.
We also face this challenge in our experiments, since collecting a large offline dataset before running an online fine-tuning experiment is difficult to accomplish within the maximum duration of a user study for a given participant.
Hence, we collect a large offline dataset of 1200 episodes of an expert user (the first author) performing tasks using their eye gaze and the default directional interface.
We use this dataset to pre-train an interface for each of the 12 participants.

\subsection{Can we learn an effective interface through RL?} \label{user-study-1}

\noindent\textbf{Offline pre-training enables efficient online fine-tuning.}
The results in Fig. \ref{fig:nav-results}a show that, through offline pre-training and online fine-tuning, ORBIT (orange) learns an interface that better enables users to navigate than the default, directional interface (gray).
For the 12 participants, ORBIT (solid orange) outperforms the directional interface (solid gray) after 100 episodes, or 30 minutes, of online fine-tuning.
In contrast, without any offline pre-training, we find that purely online RL from sparse rewards is not capable of learning an effective interface at all, even when we simulate idealized user commands (see the ablation experiment for \textbf{Q10} in Sec. \ref{ablations}).
These results show that offline pre-training is essential for making online fine-tuning from sparse rewards feasible and efficient. \looseness=-1

\noindent\textbf{ORBIT performance improves under distributional shift in command signals between the offline and online data.}
If we evaluate ORBIT on the same expert user who generated the offline data (dashed curves in Fig. \ref{fig:nav-results}a), we find that it takes less than 20 episodes of online fine-tuning before ORBIT (dashed orange) starts to outperform the directional interface (dashed gray), and that ORBIT converges to a 100\% success rate.
Furthermore, Fig. \ref{fig:nav-results}c shows that participants whose command signals are closer to the offline data distribution perform better with ORBIT.
These results suggest that ORBIT works best when the user who is operating the interface during online fine-tuning is similar to the user that generated the offline data.

While data personalization poses a potential challenge here with the discrepancy between the user who generates the offline data and the user who operates the interface, we find that offline pre-training on one user's data can still provide a good initialization for a different user, as evidenced by fine-tuning quickly developing an effective interface (solid orange in Fig. \ref{fig:nav-results}a). In spite of the data personalization challenge in our evaluation, ORBIT still outperforms the default interface after fine-tuning.
This suggests that ORBIT can be deployed to new users in practice by leveraging data from past users.
We hypothesize that with greater data diversity from multiple users, data personalization is less likely to be problematic.

\noindent\textbf{ORBIT is most helpful for users who struggle with their existing interface.}
While Fig. \ref{fig:nav-results}a shows average performance across the 12 participants, the scatter plot in Fig. \ref{fig:nav-results}b breaks down the performance of ORBIT (y-axis) vs. the directional interface (x-axis) for each participant.
We find that ORBIT tends to perform better for users who perform worse with the directional interface, and that users who can already use the directional interface well can see a small drop in performance.
These results suggest that ORBIT would be more helpful for users who struggle to use their existing interface.

\begin{figure}[t]
\vspace{0.05in}
  \begin{center}
    \includegraphics[width=\linewidth]{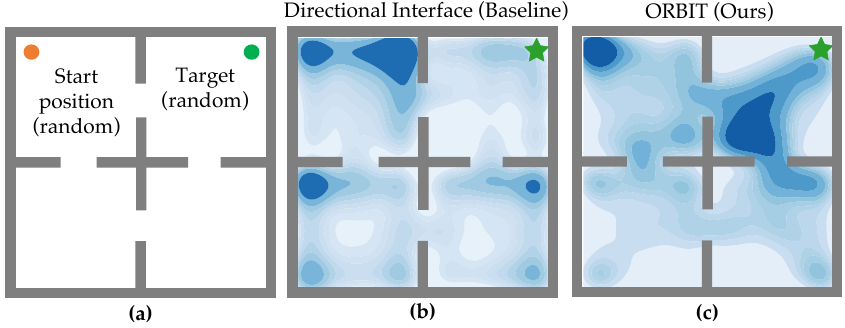}
    \vspace{-0.2in}
    \caption{\textbf{(a)} An illustration of the simulated 2D continuous navigation environment. At the start of each episode, the start position is sampled uniformly at random, and the target is randomly sampled from the four corners. \textbf{(b-c)} Heatmaps of the states that users occupy in pursuit of a target in the top-right corner (green star), where darker indicates that the user occupied that position more frequently.}
   \label{fig:nav-heat}
   \vspace{-20pt}
  \end{center}
\end{figure}

\noindent\textbf{Emergence of shared autonomy.}
Fig. \ref{fig:nav-heat}b and \ref{fig:nav-heat}c illustrate the positions that users occupy when navigating to the green star in the top-right corner.
With the directional interface, users tend to get stuck in corners of the other rooms.
With ORBIT, users spend less time in the wrong rooms, and only tend to get stuck briefly at corners or bottlenecks near the target.
These figures illustrate a key advantage of ORBIT: because the interface is trained through RL, it more often takes actions that have a chance of performing a task, while avoiding actions that are unlikely to do so.
For example, users tend to have trouble passing through the narrow corridors separating the rooms using the directional interface, since this requires precise gaze commands.
With ORBIT, the interface gets stuck less often, even with noisy gaze commands.
This ``shared autonomy'' emerges naturally through RL, without being explicitly incorporated into the interface.

\section{Ablations and Showcases in Different Domains} \label{ablations}

To perform ablations of ORBIT at a scale that would be impractical for a user study, we simulate noisy, high-dimensional user commands $\bx_t$ for the navigation task by randomly projecting the optimal 2D velocity into a 128-dimensional space of command signals, and adding i.i.d. Gaussian noise ($\sigma = 1$) to the intended velocity at each timestep. Also, similarly as in \cite{reddy2018shared}, we repeat the previous command with probability 0.75 (i.e., lagging).

We aim to answer the following.
\textbf{Q1}: Is offline RL beneficial in our setting, or can we simply use behavior cloning (BC)?
\textbf{Q2}: What if we only run BC on successful trajectories?
\textbf{Q3}: Can we directly apply vanilla IQL by simply concatenating command signals with environment states, as done in \cite{reddy2018shared}?
\textbf{Q4}: What is the effect of regularizing the trajectory encoder with negative data augmentation?
\textbf{Q5}: What is the effect of regularizing the encoders with an information bottleneck?
\textbf{Q6}: What is the effect of training a separate history encoder for the policy, compared to reusing the trajectory encoder?
\textbf{Q7}: What is the effect of conditioning the value functions on an embedding of the full trajectory vs. a partial trajectory?
\textbf{Q8}: What is the effect of training the trajectory encoder on reward prediction, instead of end-to-end training on the IQL losses $L_Q$ and $L_V$?
\textbf{Q9}: What is the effect of the offline dataset size? 
\textbf{Q10}: What if we do not have offline data, and have to acquire an interface purely through online RL?

To answer \textbf{Q1-9}, we train offline on 1K trajectories with no fine-tuning.
To answer \textbf{Q10}, we run online RL for 500 episodes with no offline pre-training.
We note that our baseline for \textbf{Q3}, where we concatenate environment states with command signals as the RL observation space, and then run offline RL directly, is to our knowledge the only type of RL baseline that can be applied to this setting. Other prior RL methods would require additional assumptions, such as a task-agnostic reward function, the user's task distribution, or a prior interface \cite{reddy2018shared,x2t2021,asha2022}. In fact, this baseline can be viewed as an offline version of \cite{reddy2018shared} without the components that would require additional assumptions.
The results in Table \ref{tab:ablation-results} show that ORBIT outperforms all of its ablated variants and the baselines. \looseness=-1

\begin{table}
\vspace{0.05in}
\begin{center}
\caption{Ablations with Simulated User Commands for Navigation}
    \begin{adjustbox}{width=\columnwidth}
    \begin{tabular}{p{5.35cm}p{1.3cm}p{1.0cm}}
    \toprule
    & Success Rate & Episode Length \\
    \midrule
    Directional Interface (Baseline) & $ 79.9 \pm 5.8 $ & $98 \pm 8$ \\
    Behavior Cloning (Baseline) (\textbf{Q1}) & $ 72.5 \pm 0.3 $ & $ 117 \pm 1 $ \\
    Filtered Behavior Cloning (Baseline) (\textbf{Q2}) & $ 90.0 \pm 0.2 $ & $ 95 \pm 1 $  \\
    Vanilla IQL (Baseline based on \cite{reddy2018shared}) (\textbf{Q3}) & $28.5 \pm 0.3 $ & $147 \pm 1 $ 
    \\
    \midrule
    \textbf{ORBIT (Ours)} & $\mathbf{ 95.2 \pm 1.0 }$ & $\mathbf{ 66 \pm 2 }$ \\
    ORBIT w/o NDA (\textbf{Q4}) & $ 29.7 \pm 1.3 $ & $ 146 \pm 2 $ \\
    ORBIT w/o VIB (\textbf{Q5}) & $ 35.9 \pm 3.9 $ & $ 139 \pm 4 $ \\
    ORBIT w/o Separate History Encoder (\textbf{Q6}) & $ 68.9 \pm 1.9 $ & $ 95 \pm 4 $ \\
    ORBIT w/ Partial Trajectory Embedding (\textbf{Q7}) & $ 66.9 \pm 3.8 $ & $ 106 \pm 4 $ \\
    ORBIT w/ End-to-End Trajectory Encoder (\textbf{Q8}) & $ 27.9 \pm 0.6 $ & $ 148 \pm 1 $ \\
    ORBIT w/ 500 Offline Trajectories (\textbf{Q9}) & $ 89.5 \pm 2.1 $ & $ 77 \pm 3 $ \\
    ORBIT w/ 100 Offline Trajectories (\textbf{Q9}) & $ 69.3 \pm 11.7 $ & $ 103 \pm 11 $ \\
    ORBIT w/o Offline Pre-Training (\textbf{Q10}) & $ 17.9 \pm 0.3 $ & $ 176 \pm 1 $ \\
    \bottomrule
  \end{tabular}
  \end{adjustbox}
   \label{tab:ablation-results}
   \caption*{Evaluations on 500 episodes and 3 random seeds}
    \vspace{-30pt}
   \end{center}
\end{table}

\begin{table}[t]
\begin{center}
\vspace{0.05in}
\caption{Lunar Lander and Sawyer Pushing Experiments}
    \begin{adjustbox}{width=\columnwidth}
    \begin{tabular}{lll}
    \toprule
    &
    \includegraphics[width=0.25\linewidth]{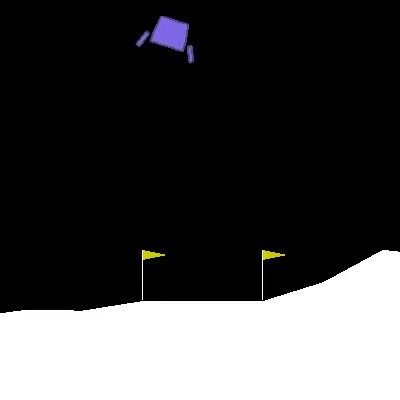} & \includegraphics[width=0.25\linewidth]{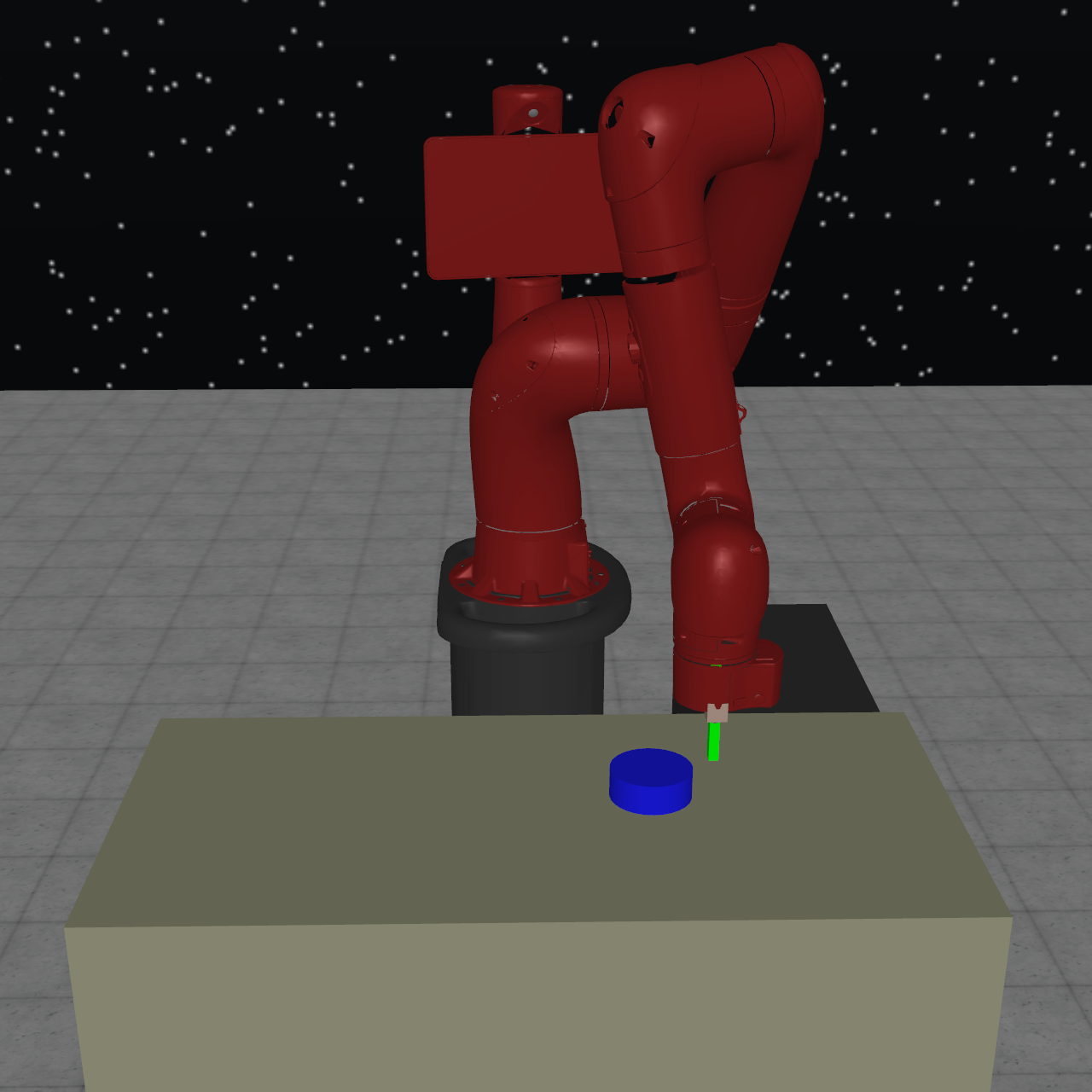} \\
    & (Simulated Commands) & (Webcam Commands) \\
    & Success Rate & Success Rate \\
    Directional & $ 66.9 \pm 1.8 $ & $57.9 \pm 4.8$  \\
    \hline
    \textbf{ORBIT} & $\mathbf{ 85.5 \pm 1.1}$ & $\mathbf{73}$\\
    \bottomrule
    \end{tabular}
    \end{adjustbox}
    \label{tab:sawyer-lander-results}
   \caption*{Lunar Lander: evaluations on 500 episodes and 3 random seeds. Sawyer Pushing: evaluations on 100 episodes and 1 random seed.}
   \vspace{-20pt}
\end{center}
\end{table}
To showcase ORBIT in other domains, we evaluate ORBIT on the Lunar Lander and Sawyer Pushing tasks from \cite{ghosh2019learning} (images in Table \ref{tab:sawyer-lander-results}). 
For Sawyer Pushing, we have an expert user (the first author) provide real webcam command signals.
For Lunar Lander, we simulate noisy, high-dimensional user commands as in the ablations.
The results in Table \ref{tab:sawyer-lander-results} show that ORBIT outperforms the default interface in both domains.

\section{Discussion} \label{discussion}

Through a user study with 12 participants, we show that offline RL can be used to bootstrap the online fine-tuning of an assistive human-machine interface that enables people to perform a simulated navigation task by using their eye gaze to modulate a noisy, 128-dimensional command signal from their webcam.
Large-scale ablation experiments with simulated user commands show that, while our method has many moving parts, each is well-motivated and enables the user to succeed at their desired tasks more frequently. \looseness=-1

Our experiments are limited to relatively simple tasks in simulated domains. Scaling an offline RL method such as ORBIT to more practical and challenging real-world applications may require larger and more diverse datasets. This could be achieved through more widespread use of assistive control systems, such that users are constantly collecting data while performing daily activities, and data may be shared across users.

ORBIT is limited by its inability to learn from data collected in the environment without the user in the loop.
In real-world applications such as assistive robotics, the robot can interact with the environment in an unsupervised manner without the user present, e.g., in order to explore the state space or learn the environment dynamics.
A promising direction for future work would be to integrate data that is collected autonomously into the ORBIT pipeline, using an algorithm like COG \cite{singh2020cog}.
Another promising idea is to use offline meta-RL \cite{pong2021offline,zhao2021offline,mitchell2021offline} to learn from heterogeneous offline data collected from multiple users, or from the same user at different times. \looseness=-1

\section{Acknowledgements} \label{acknowledgements}
Thanks to members of the InterACT and RAIL labs at UC Berkeley for feedback on this project, especially Sean Chen. This work was supported by ARL DCIST CRA W911NF-17-2-0181, ARO W911NF-21-1-0097, Weill Neurohub, Semiconductor Research Corporation, and CIFAR.

\bibliographystyle{unsrt}
\bibliography{main}

\clearpage

\appendix

\section{Appendix}
\subsection{History Encoder Architecture} \label{hist-enc-arch}

When choosing an architecture for the history encoder $e_{\eta}$, we initially tried the same recurrent architecture as the trajectory encoder from Sec. \ref{learn-value}, but found that this tended to overfit to the sequences of states in the offline data---see the ablation experiment for $\textbf{Q6}$ in Sec. \ref{ablations}.
To address this issue, we model the history encoder $e_{\eta}$ as a product of independent Gaussian factors, similar to the inference network architecture used by \cite{rakelly2019efficient}.
This results in a Gaussian posterior,
\begin{equation}
\small
    e_{\eta}(\bz_t | h_{0:t}) \propto \prod_{i=0}^t \mathcal{N} (\bz_t | f_{\eta}^{\mu}(\bs_i, \bx_i), f_{\eta}^{\sigma}(\bs_i, \bx_i)).
\end{equation}
This architecture requires fewer parameters $\eta$ than an equivalent recurrent architecture, but is invariant to the order of the timesteps.

\subsection{Implementation} \label{imp-details}

For the navigation task, we pre-train the trajectory encoder $e_{\phi}$ and history encoder $e_{\eta}$ with 50K gradient steps, the value functions $Q$ and $V$ with 200K steps, and the policy $\pi_{\theta}$ with 200K steps.
For the pushing task, we pre-train the encoders with 100K gradient steps, and the value functions with 300K steps.
We set $\beta_1 = 0.005$ in Equation \ref{eqn:val-seq-enc-loss} for the navigation task, $\beta_1 = 0.05$ for the pushing task, and $\beta_1 = 0.001$ for the Lunar Lander game.
We set $\beta_3 = 0.1$ in Equation \ref{eqn:pol-loss} for the navigation task in the user study, $\beta_3 = 0.02$ for the pushing task, and $\beta_3 = 0.01$ for the navigation task in the ablation experiments as well as the Lunar Lander game simulations.
In IQL, we set the expectile $\tau = 0.8$.
We set $\beta_2 = 1/3$ in Equation \ref{eqn:weighted-bc} in the navigation task, $\beta_2 = 0.5$ for the pushing task, and $\beta_2 = 0.02$ in the Lunar Lander game.
We set the dimensionality of the latent space to $k = 2$ in the navigation task, $k = 2$ for the pushing task, and $k = 1$ in the Lunar Lander game.
To optimize all of our losses, we use the Adam optimizer \cite{kingma2014adam} with an initial learning rate of $3 \cdot 10^{-4}$.
Following the IQL method, we use a constant learning rate for training the value functions, and cosine scheduling for training the encoders and policy.
We use a batch size of 16 sequences to train the trajectory encoder, and batch size of 256 to train the value functions and policy.
We set the soft $Q$-function update $\tau = 0.005$.
We set $\beta_4 = 1$ in line \ref{lst:line:fine-tune-traj-enc} of Alg. \ref{alg:orbit-alg}.

The trajectory encoder architecture maps $(\bs_t, \bx_t)$ $\rightarrow$ 64-dimensional linear layer $\rightarrow$ ReLU $\rightarrow$  (previous representation, $r_t$) $\rightarrow$ 64-dimensional GRU $\rightarrow$ linear layer $\rightarrow$ means and standard deviations of latent features.
For the pushing task and the Lunar Lander game, we apply a dropout rate of 0.5 to the output of the first ReLU layer, and increase the number of hidden units in the GRU layer from 64 to 512.
The reward decoder architecture is a feedforward network with 2 layers of 32 hidden units each and ReLU activations.
The history encoder architecture is the same, but with 64 hidden units, and a sliding window of the past 30 timesteps in the navigation task, the past 50 timesteps in the pushing task, and the past 200 timesteps (i.e., the full history) in the Lunar Lander game.
For the pushing task and the Lunar Lander game, we increase the number of hidden units in the second hidden layer from 64 to 512.
The $Q$-function and value function architectures are both feedforward networks with 2 layers of 64 hidden units each and ReLU activations, and a dropout rate of 0.01.
The policy architecture is a feedforward network with 2 layers of 64 hidden units each and ReLU activations, with a dropout rate of 0.1.
For the pushing task and the Lunar Lander game, we increase the number of hidden units in both hidden layers of the $Q$-function, value function, and policy architectures from 64 to 256.

For the navigation task, for each gradient step on the trajectory encoder loss in Equation \ref{eqn:val-seq-enc-loss}, we compute the loss terms $\ell_{\mathrm{RPE}}$ and $\ell_{\mathrm{NDA}}$ as follows.
We sample a batch of trajectories, compute the mean of the loss terms associated with positive rewards in the binary cross-entropy loss $\ell_{\mathrm{RPE}}$, and compute the mean of the loss terms associated with negative rewards.
We then compute the negative data augmentation loss $\ell_{\mathrm{NDA}}$ for the batch by randomly permuting the alignment between embeddings and state sequences, and taking the mean of the loss terms.
We then weigh the three mean loss terms---positive rewards, negative rewards, and negative data augmentation---equally when taking a gradient step.
For the Lunar Lander and pushing tasks, we take a slightly different approach to weighting these terms.
The positive rewards have a weight of $\frac{1}{2}$, negative rewards a weight of $\frac{1}{6}$, negative data augmentation a weight of $\frac{1}{6}$, and negative data augmentation for transitions with a reward of $+1$ before permutation.

For all offline pre-training and simulation experiments, we trained and ran our models on an RTX 2080 GPU.
We performed the user studies on one consumer-grade desktop computer with a Logitech C920x webcam.

Where applicable, and if not otherwise specified, our implementation of baseline methods and ablations of ORBIT use the same hyperparameter values as our implementation of the full ORBIT algorithm.

\subsection{Simulation Experiments} \label{sim-details}

When simulating noisy user commands, we add Gaussian noise with standard deviation $\sigma = 1$ in the navigation task, and $\sigma = 0.25$ in the Lunar Lander game.
We repeat the user's previous command (i.e., lag) with probability $0.75$ in the navigation task, and $0.25$ in the Lunar Lander game.
To follow the real eye gaze capture pipeline from the user study as closely as possible, we apply a random linear projection to the simulated user's noisy 2D velocity that transforms this 2D intent into a 128-dimensional command signal, then clamp the projected signal using the ReLU function.
When calibrating the default directional interface, we only collect 10 samples per direction (in contrast to the 20 samples used in the user study).
The offline data consists of 100 episodes per default interface---10 different default interfaces (each trained with a different random seed) in the navigation task, and 50 different default interfaces in the Lunar Lander game---yielding a total of 1K episodes (98253 timesteps) in the navigation task, and 5K episodes (474984 timesteps) in the Lunar Lander game.

We train an agent to act as the simulated user using the proximal policy optimization algorithm (PPO) \cite{schulman2017proximal} and the default hyperparameter values from the Stable Baselines3 library \cite{raffin2021stable}.
We take 2048 environment steps per gradient step in the navigation task, and 8192 environment steps per gradient step in the Lunar Lander game.
We set the entropy coefficient to 0.01.
The policy architecture is a feedforward network with 2 layers of 32 hidden units each and ReLU activations.
We run PPO for 3 million timesteps in the navigation task, and 1M timesteps in the Lunar Lander game.
We sample actions from the policy, instead of executing the action with the highest likelihood according to the policy.

For the ablation experiment addressing \textbf{Q8}, we set $\beta_1 = 0.05$ in Equation \ref{eqn:val-seq-enc-loss}.

For the ablation experiment addressing \textbf{Q10}, we run online RL from scratch with no offline pre-training.
Unlike the online fine-tuning procedure described in Section \ref{fine-tuning}, for this experiment we use negative data augmentation and do not freeze the reward decoder.
We take the same number of gradient steps per episode as the number of environment steps in the previous episode.
We set $\beta_4 = 0$ in line \ref{lst:line:fine-tune-traj-enc} of Alg. \ref{alg:orbit-alg}.

For the Lunar Lander simulation experiments, we collect an offline dataset by calibrating 20 different default interfaces and collecting 100 episodes using each of them, yielding a total of 2K episodes total in the offline dataset.

\subsection{Sawyer Pushing Experiments with Real Webcam Commands}

We found that online fine-tuning did not substantially improve performance, so Table \ref{tab:sawyer-lander-results} shows the results for purely-offline RL (i.e., no online fine-tuning).
We collect the offline dataset by calibrating 12 different default interfaces and collecting 100 episodes using each of them, yielding a total of 1200 episodes (180561 timesteps).

In our initial experiments with the pushing task, we found that the trajectory encoder learned to generate latent embeddings that, instead of representing the user's intent, represent whether or not the states in the trajectory are associated with high or low reward---this leads to the formation of two separate clusters of latents for successes and failures, instead of separate clusters for different goals or tasks.
To address this issue, we train a discriminator to classify whether a latent embedding belongs to a successful trajectory or failed trajectory.
We class-balance this binary classification loss to contain equal proportions of positive and negative examples.
We then add the following term to the trajectory encoder training loss in Equation \ref{eqn:val-seq-enc-loss}: the log-likelihood of the discriminator classifying latents from failed trajectories as successful.
This additional term encourages the distribution of latent embeddings of successful trajectories to match the distribution of latents for failed trajectories.
We alternate between taking 1 gradient step on the binary classification loss to update the discriminator, and taking 1 gradient step on Equation \ref{eqn:val-seq-enc-loss}.
We weight both loss terms with a coefficient of $0.5$.
We model the discriminator as a feedforward network with 2 hidden layers of 32 units each and ReLU activations.
To determine whether the discriminator loss could be useful for a particular domain, we recommend visualizing the latent embeddings of the user's intent---if there are two distinct clusters corresponding to successes and failures, then the discriminator loss could improve intent inference.

\subsection{Environments} \label{env-details}

In the pushing task, there are 4 possible 2D goal positions for the puck, each in the corner of a rectangle.
At the beginning of each episode, the puck and end effector are both reset to the center of this rectangle.
There are 8 discrete actions, corresponding to unit velocities in the 4 cardinal directions and 4 diagonal directions.
The 4D state consists of the 2D end effector position and 2D puck position.
The state and goal positions are normalized to the range $[-1, 1]$.

In the Lunar Lander game, we flatten the terrain, partition the terrain into 5 equally-sized landing zones, and choose one uniformly at random at the beginning of each episode to be the simulated user's desired landing zone.
To make the game easier, we set the initial force perturbation on the lander to zero.

The maximum episode length in the navigation task, pushing task, and the Lunar Lander game is 200 timesteps.

Sparse rewards are set to $-1$ and $0$ in the navigation and pushing tasks (to encourage faster task completion), and $0$ and $1$ in the Lunar Lander game (since episode length does not matter).
The value functions in the Lunar Lander game have a sigmoid output activation layer, since the values are bounded between $0$ and $1$.

\subsection{User Study} \label{study-details}

\begin{figure*}[t]
    \centering
    \includegraphics[width=\linewidth]{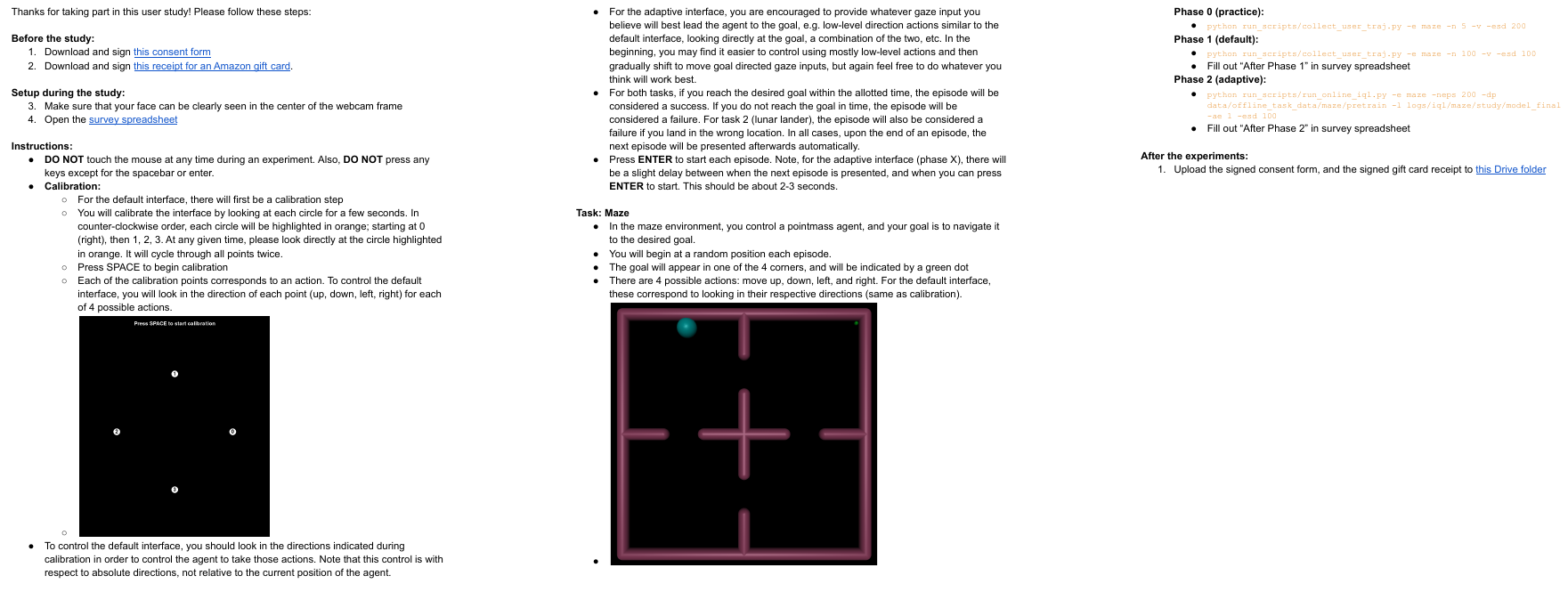}
    \caption{Instructions for user study participants}
    \label{fig:protocol}
\end{figure*}

\begin{table*}[t]
  \caption{Subjective Evaluations from User Study Participants}
  \centering
\small
\begin{tabular}{llll}
\toprule
{} &   $p$ & Directional Interface & ORBIT\\
\midrule
The system performed the task I wanted & $>.05$ & 4.42 & 4.75 \\
I felt in control & $>.05$ & 4.50 & 4.17 \\
The system responded to my input in the way that I expected & $>.05$ & 4.33 & 3.67 \\
The system was competent at performing tasks... & & & \\
...even if they weren't the tasks I wanted & $>.05$ & 3.92 & 4.83 \\
\textbf{The system improved over time} & $\mathbf{<.0001}$ & 2.50 & \textbf{6.17} \\
I improved at using the system over time & $>.05$ & 4.92 & 5.50 \\
I always looked directly at my final target... & & & \\
...holding the same gaze throughout an episode & $>.05$ & 1.75 & 2.75 \\
I compensated for flaws in the system... & & & \\
...by changing my gaze over time... & & & \\
..., e.g., by exaggerating looking in certain directions & $>.05$ & 5.00 & 5.55 \\
\bottomrule
\end{tabular}
  \caption*{Means reported for responses on a 7-point Likert scale, where 1 = Strongly Disagree, 4 = Neither Disagree nor Agree, and 7 = Strongly Agree. $p$-values from a one-way repeated measures ANOVA with the presence of ORBIT as a factor influencing responses.}
  \label{tab:user-study-survey}
\end{table*}

We recruited 9 male and 3 female participants, with an average age of 22.
We obtained informed consent from each participant, as well as IRB approval for our study.
Each participant was provided with the rules of the task (see Fig. \ref{fig:protocol}), and compensated with a \$20 Amazon gift card.

Given that the reward for the navigation task is very simple for a person to compute (+1 for reaching the goal, 0 otherwise), we automate the reward signal instead of asking the user for a button press.
In a similar user study from prior work \cite{x2t2021}, it has been shown that automated sparse rewards match user-provided rewards 98.6\% of the time.

All participants used the same webcam setup and the same desktop computer in the same room.
The offline data was collected from the expert user (the first author) in 12 sessions of 100 episodes (94100 steps total), with a new default directional interface calibrated at the beginning of each session to account for changes in the webcam image distribution (e.g., background lighting).
During the online fine-tuning phase, we take $n = 20$ and $m = 20$ gradient steps in lines \ref{lst:line:n-grad} and \ref{lst:line:m-grad} in Alg. \ref{alg:orbit-alg}.
We use the Adam optimizer with the same hyperparameter settings as in the offline pre-training phase (i.e., a constant learning rate of $3 \cdot 10^{-4}$).

To capture the user's eye gaze command signals $\bx_t$, we used code from prior work \cite{x2t2021,asha2022,krafka2016eye}.

The only personal data logged during the user studies were 128-dimensional eye gaze features, which cannot be used to uniquely recover the original face or eye images from the webcam.

When prompted to ``please describe your command strategy'', participants responded as follows.
\begin{displayquote}
\small
User 5: \\
After Phase A: \\
\textbf{Feel like it's much more difficult to go left-right than top-down. Probably because the face shifts more in the horizontal direction. So every time I feel that it is very very difficult say to go right, I'll shift face leftwards and it becomes easier, vice versa.} \\
After Phase B: \\
\textbf{Look at final target if in same quadrant, otherwise look at some waypoints} \\
\\
User 6: \\
After Phase A: \\
\textbf{I generally tried to look to the edge of the screen, but would sometimes follow the dot in different directions.} \\
After Phase B: \\
\textbf{First I would look at the target, and if the ball got stuck, I would look at the exaggerated direction of the target, or use basic left/right/up/down directions.} \\
\\
User 7: \\
After Phase A: \\
\textbf{I'd move my eyes in the direction I wanted the ball to move} \\
\\
User 8: \\
After Phase A: \\
\textbf{To get through door: 
first get on wall of door. Then gaze in one direction and quickly switch gaze direction to try to get through door. Repeat until through door 
If in same room as target: 
first gaze in one direction to get on on wall such that i am strraight line away from target. Then move to direct by gazing in one direction} \\
After Phase B: \\
\textbf{Look at either center of rooms or the target location if already in same room as target location. Do exaggerated eye motions to get unstuck if stuck on a wall.} \\
\\
User 11: \\
After Phase A: \\
\textbf{At first I looked  at the points shown during calibration, but later I found there were particular points on the screen which more reliably mapped to the direction I wanted (even if they weren't exactly the calibration points), so I looekd at those.} \\
After Phase B: \\
\textbf{Look at either the goal or the nearest doorway between rooms. If the agent isn't responding, alternate strategies.} \\
\\
User 12:\\
After Phase A: \\
\textbf{Mostly looking directly at the direction I wanted the agent to move in at a finer-grained level}\\
After Phase B: \\
\textbf{I looked directly at the goal initially, but if that didn't work or the agent got stuck, then I would look in the direction I wanted the agent to move in (up/down/left/right) at a lower level}
\end{displayquote}

The heatmaps in Fig. \ref{fig:nav-heat} are generated by filtering out state transitions where consecutives states are identical to each other (i.e., repeated).
For ORBIT, we only visualize the last 50 episodes of fine-tuning.

\end{document}